\pdfoutput=1

\documentclass[11pt]{article}
\usepackage[table]{xcolor}
\usepackage[final]{acl}

\usepackage{times}
\usepackage{latexsym}

\usepackage[T1]{fontenc}

\usepackage[utf8]{inputenc}

\usepackage{microtype}

\usepackage{inconsolata}

\usepackage{graphicx}

\usepackage[normalem]{ulem}
\usepackage{CJKutf8}
\usepackage{booktabs}
\usepackage{comment}

\usepackage{subcaption}
\usepackage{tabularx}
\usepackage{float}
\usepackage{afterpage}
\usepackage{gb4e}
\usepackage{ulem}
\noautomath

\title{Who Relies More on World Knowledge and Bias for Syntactic Ambiguity Resolution: Humans or LLMs?}

\author{
  \textbf{So Young Lee\textsuperscript{$\dagger$}},
  \textbf{Russell Scheinberg\textsuperscript{$\diamondsuit$}},
  \textbf{Amber Shore\textsuperscript{$\diamondsuit$}},
  \textbf{Ameeta Agrawal\textsuperscript{$\diamondsuit$}}
\\
  \textsuperscript{$\dagger$}Miami University, USA\\
  \textsuperscript{$\diamondsuit$}Portland State University, USA
\\
\texttt{soyoung.lee@miamioh.edu}\\\texttt{\{rschein2,ashore,ameeta\}@pdx.edu}
}

\definecolor{mycustomcolor}{RGB}{74, 0, 255}

\begin{document}
\maketitle
\begin{abstract}
This study explores how recent large language models (LLMs) navigate relative clause attachment {ambiguity} and use world knowledge biases for disambiguation in six typologically diverse languages: English, Chinese, Japanese, Korean, Russian, and Spanish. We describe the process of creating a novel dataset -- \texttt{MultiWho}\footnote{Dataset available at \url{https://github.com/PortNLP/MultiWHO}.} -- for fine-grained evaluation of relative clause attachment preferences in ambiguous and unambiguous contexts. Our experiments with three LLMs indicate that, contrary to humans, LLMs consistently exhibit a preference for local attachment, displaying limited responsiveness to syntactic variations or language-specific attachment patterns. Although LLMs performed well in unambiguous cases, they rigidly prioritized world knowledge biases, lacking the flexibility of human language processing. These findings highlight the need for more diverse, pragmatically nuanced multilingual training to improve LLMs' handling of complex structures and human-like comprehension.

\end{abstract}

\section{Introduction}
Natural language is inherently ambiguous, with single expressions often having multiple interpretations. This ambiguity poses significant challenges to both human cognition and computational models, especially in tasks requiring precise language understanding like machine translation, question answering, and dialogue systems. Miscommunication can arise when different listeners or readers interpret the same expression differently, making ambiguity resolution a critical area of research \cite{mehrabi-etal-2023-resolving,he2024enhancing,niwa2024ambignlgaddressingtaskambiguity, hatami-etal-2022-analysing, tran-etal-2022-multimodal, futeral-etal-2023-tackling, nath-etal-2024-multimodal}. 


\begin{figure}[t!]
    \centering
    \includegraphics[width=1\linewidth]{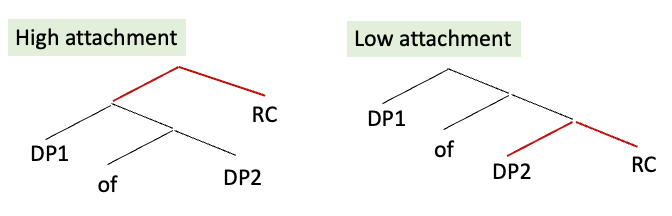}
    \caption{Syntactic Structures of DP1 Modification (left) and DP2 Modification (right) in English}
    \label{fig:hala} 
\end{figure}

Among various types of ambiguity, this study focuses on syntactic ambiguity, specifically relative clause (RC) attachment ambiguity. Syntactic ambiguity occurs when a sentence's structure allows for multiple grammatical interpretations. RC attachment ambiguity arises when a relative clause can attach to more than one determiner phrase (DP), leading to different possible meanings. For instance, in (\ref{eg:rc1}), the RC \textit{who had a beard} could refer to either the local DP (DP2) \textit{the man} or the non-local DP (DP1) \textit{the son}. 

\begin{figure*}[t!]
    \centering
\includegraphics[width=0.99\textwidth,trim=0 25 0 0, clip]{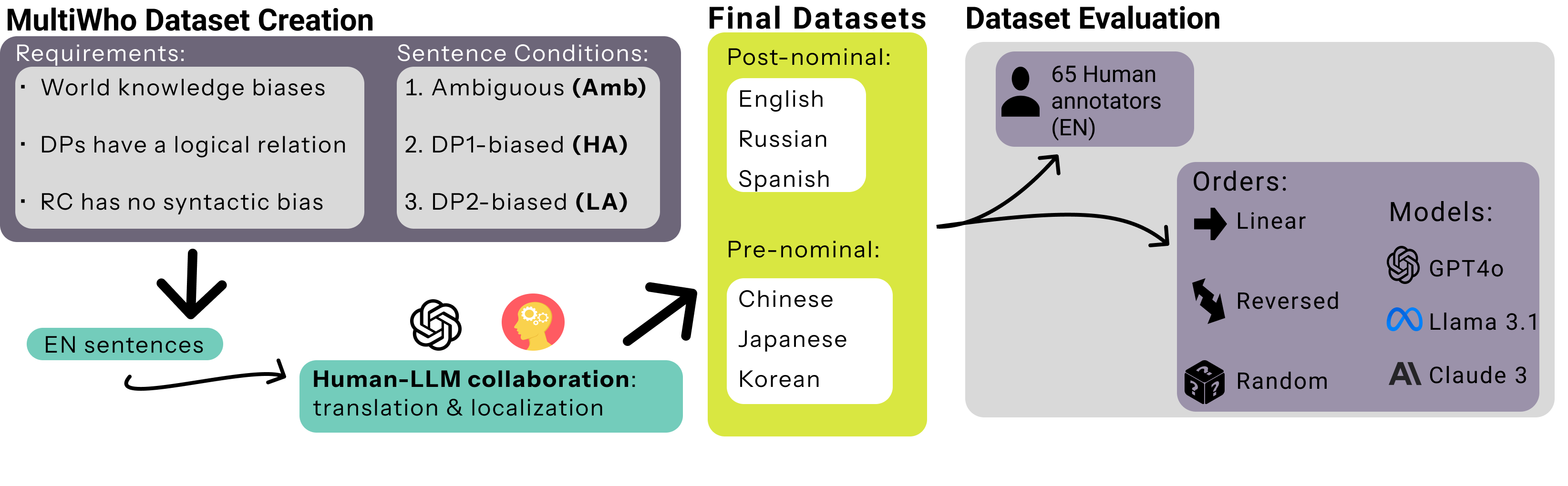}
    \caption{\textbf{MultiWho Dataset:} The dataset creation started with a list of \textbf{requirements} and three different \textbf{conditions}. 
    Using a collaborative \textbf{human-LLM} process, we started with developing English sentences and continued through translation and localization, resulting in a multilingual dataset across \textbf{six languages}.
    While not all sentences are pragmatically equivalent in all languages, they are structurally equivalent with regard to our requirements.
    These datasets were \textbf{evaluated} in two ways: the English dataset was evaluated by 65 human annotators for ambiguity/DP-bias, and all 6 datasets were evaluated for ambiguity/DP-bias in three different answer order settings by LLMs.
    }
    \label{fig:dataset_overview}
\end{figure*}

\begin{exe}
\ex The doctor met the son$_{DP1}$ of the man$_{DP2}$ [who had a beard]$_{RC}$. 
\begin{xlist}
        \ex \textbf{DP2 modification} (linearly local, structurally low attachment (LA)): The person who had a beard is the man.
        \ex \textbf{DP1 modification} (linearly non-local, structurally high attachment (HA)): The person who had a beard is the son.
    \end{xlist}
\label{eg:rc1}
\end{exe}

Figure \ref{fig:hala} illustrates this syntactic ambiguity with two distinct syntax trees. Previous studies have demonstrated that attachment preferences are language-specific, influenced by multiple factors in the resolution of syntactic ambiguities \cite{clifton2003use, fodor1998learning, grillo2014novel, hemforth1996syntactic, traxler1998adjunct}. While world knowledge and biases often guide interpretation, the extent to which these factors override inherent syntactic parsing preferences remains underexplored — especially in large language models (LLMs) \citep{openai2024gpt4technicalreport, touvron2023llamaopenefficientfoundation}. Moreover, much of the research on LLMs and ambiguity resolution has focused on English and other Indo-European languages, leaving languages with different syntactic structures less examined \cite{yuan-etal-2023-ambicoref,cai2024largelanguagemodelsresemble}.

This study explores how humans and LLMs resolve RC attachment ambiguities across languages and whether world knowledge overrides syntactic preferences.\footnote{As defined in the previous studies \citep{kecskes2023interplay,sayeedexplicit}, world knowledge and bias, as used in our study, refer to societal norms, plausibility, and stereotypes. These factors shape predictions but are distinct from semantic cues, which are usually narrowly defined in linguistic contexts (e.g., animacy or thematic roles).} We investigate if LLMs show consistent attachment patterns across languages and how they align with human processing. We selected six languages—English (EN), Chinese (ZH), Japanese (JP), Korean (KO), Russian (RU), and Spanish (ES)—for their syntactic diversity, particularly regarding the position of RCs. For instance, EN, ES, and RU use postnominal RCs that follow the DP they modify, as shown in (\ref{eg:postnominal}):

\begin{exe}
    \ex DP RC: (English)\\ the man [who ran the marathon]
    \label{eg:postnominal}
\end{exe}

In contrast, JP, KO, and ZH use prenominal RCs that precede the DP they modify, as illustrated in the KO example (\ref{k1-com}):

\begin{exe}
\ex RC DP: (Korean)\label{k1-com}\vspace{-0.3cm}
\gll [\begin{CJK}{UTF8}{mj}마라톤을\end{CJK} \begin{CJK}{UTF8}{mj}뛴\end{CJK}] \begin{CJK}{UTF8}{mj}남자\end{CJK}\\ 
 [marathon ran] man\\
\end{exe}

Figure~\ref{fig:dataset_overview} presents a schematic figure of the components of this study. Our research addresses the following questions.
\begin{enumerate}
\itemsep-0.4em
\item What are the syntactic preferences of LLMs to resolve the ambiguities in RC attachment in different languages? Will there be a certain pattern depending on the syntactic differences among languages?
\item How does the sensitivity of LLMs to world knowledge and biases compare to that of humans when resolving syntactic ambiguities?
    \item Does the presentation order (linear, reversed, random) of the possible response choices (DP1 or DP2) influence LLM attachment preferences?
\end{enumerate}

\setlength{\tabcolsep}{12pt}
\begin{table*}[!t]
\centering
\begin{tabularx}{\linewidth}{c l}
\hline
\textbf{Condition} & \textbf{Example Sentence} \\ 
\hline
Ambiguous & The doctor met \textbf{the son} of \textbf{the man} who had a beard. \\ 
DP1 Biased & The doctor met \textbf{the son} of the woman who had a beard. \\ 
DP2 Biased & The doctor met the daughter of \textbf{the man} who had a beard. \\ 
\hline
\end{tabularx} 
\caption{Example set of English stimuli for RC attachment ambiguity from \texttt{MultiWho} dataset. In the ambiguous condition, the absence of clear world knowledge and biases allows both DP1 \textit{the son} and DP2 \textit{the man} to serve as equally plausible referents for the RC. In the DP1-biased condition, the less plausible scenario of a woman having a beard leads to a preference for attaching the relative clause to DP1 \textit{the son}. In contrast, the DP2-biased condition exploits the greater plausibility of a man having a beard, which favors attachment to DP2 \textit{the man}.}
\label{tab:example_stimuli} 
\end{table*}

To briefly summarize our results, in ambiguous cases in English, we found that humans exhibit a strong Low Attachment (LA) preference, with an HA (High Attachment) response rate of approximately 0.2, while LLMs show an even stronger LA preference of around 0.1. Extending this observation to other languages, LLMs default to LA preference as well, aligning with human preferences in EN and ZH but contrasting with the HA preferences observed in humans for KO, JP, RU, and ES. In unambiguous EN cases, humans demonstrate a stronger and more consistent adherence to DP2-bias (0.88) compared to DP1-bias (0.64). This reflects a natural tendency toward LA structures and showcases their flexibility in adjusting interpretations despite inconsistencies in world knowledge. In contrast, LLMs display near-perfect accuracy for both DP2- and DP1-biases, suggesting a greater sensitivity to world knowledge biases and a rigidity in interpretations that may reinforce existing social stereotypes. Similarly, in unambiguous cases across other languages, LLMs show high accuracy for both biases, with higher accuracy for DP2, mirroring the patterns observed in humans in EN. 

This study's main contributions are threefold: (1) we create a new dataset for assessing the performance of LLMs in resolving ambiguities across multiple languages, (2) we describe the iterative process of linguist-LLM collaboration for generating such a dataset, and (3) we analyze how LLM performance compares to human processing patterns, identifying the limitations of LLMs and providing insights for future improvements. 



\section{\texttt{MultiWho} Dataset} 
We introduce \texttt{MultiWho}, a dataset designed to examine how {humans and} LLMs utilize world knowledge and biases to resolve syntactic ambiguities.
 \texttt{MultiWho} comprises a total of 1728 sentences, with 96 sets, spanning three categories (ambiguous, DP1-biased, and DP2-biased), in six languages, EN, ES, JP, KO, RU, and ZH.
A sample  English set is shown in Table \ref{tab:example_stimuli}.


Recently, LLMs' fluency and instruction-following ability have prompted interest in human-LLM collaboration for dataset creation \cite{long2024llmsdrivensyntheticdatageneration,10.1145/3613904.3641960}. However, even advanced LLMs such as GPT-4o and Claude 3 Sonnet were unable to {consistently} generate sentences that met our criteria. This led us to develop a more flexible paradigm for specialized dataset creation, where human experts leverage LLMs for assistance and support rather than complete solutions. Our approach involved
\textit{(i)} brainstorming with LLMs to generate options meeting individual constraints,
\textit{(ii)} manually combining these elements to create sentences satisfying all criteria,
\textit{(iii)} using LLMs for validation and discussion throughout the process,
\textit{(iv)} employing LLMs' fluency and cultural knowledge for initial drafts in multiple languages, and
\textit{(v)} validating and refining these drafts with help from native speakers.

This collaborative method allowed us to work efficiently while maintaining high standards of quality and consistency across diverse languages.

\subsection{Dataset Creation}
\label{sec:dataset}

We started in EN with the construction [DP1 of DP2 RC], where the RC could syntactically attach to DP1 (HA) or DP2 (LA) (see Figure \ref{fig:hala}). A main subject and verb are added to complete the sentence.
Importantly, we consistently use DP1 and DP2 to refer to the {\em structural positions in the syntax tree}, not the linear order of the words in the sentence, which varies by language: in postnominal languages (EN, ES, RU), the linear order \textit{matches} the structural order: DP1 DP2 RC, while in postnominal languages (ZH, JP, KO), the linear order \textit{is reversed}: RC DP2 DP1. 


Sentences were designed to fall into one of three congruency categories: 
\begin{itemize}
\itemsep-0.4em 
    \item \textbf{ambiguous}, where no clear bias cue favors one attachment over the other;
    \item \textbf{DP1-biased}, where the bias content of the sentence favors attachment to DP1; and
    \item \textbf{DP2-biased}, where world knowledge and biases favor attachment to DP2. 
\end{itemize}

\medskip

\noindent \textbf{\em Constraints for potential linguistic factors} \quad Bias is achieved by the RC applying differentially to the DPs (see Table \ref{tab:example_stimuli}).
We further enforced the following constraints. 
\textbf{Morphosyntactic constraints}: To prevent biases from word length and animacy, DPs must be single words referring to humans. Additionally, the RC should exclude any morphosyntactic markers, such as grammatical number or gender, that would require agreement with either DP (e.g. ``the \textbf{sons} of the doctor who \textbf{were} at home'').
\textbf{Semantic relation between DPs:} The \textit{DP1 of DP2} phrase must form a plausible relationship to exclude expressions like ``the firefighter of the baby''.
\textbf{Naturalness:} The subject and main verb of the sentence must be plausible in light of the rest of the sentence.

\medskip
\noindent \textbf{\em Categories of world knowledge and biases} \quad We employed various world knowledge and biases to guide RC attachment, such as {\bf gender}: physiological differences (e.g., `giving birth' for female bias) and gender roles (e.g.,  `participating in a beauty contest' in RU);  {\bf age}: e.g., `the brother of the baby who was driving a car', and {\bf profession}: e.g., `the interpreter of the judge who was monolingual'. We recognize that some of these biases are based on societal stereotypes, but we utilize them as they reflect current linguistic associations and world knowledge (also see Section \ref{sec:limitations}, Limitations).

\subsection{Iterative Linguist-LLM Collaboration}



While they can generate outputs quickly, the current generation of LLMs is unable to respect the large number of constraints simultaneously. For example, when reminded that ``the firefighter of the baby'' breaks the semantic relation constraint, it might generate ``the toy car of the baby'', breaking the single word and animacy constraints. LLMs also often accept artificial-sounding sentences, and are hampered by their demonstrated yes-response bias \cite{dentella_yes_bias}, particularly in languages other than English. The iterative process involved a linguist using the LLMs where possible while recognizing and working around its limitations. The linguist offset the weaknesses by verifying constraints, improving readability and naturalness (with help of native speakers), and correcting otherwise implausible sentences.

Although LLMs are unable to uphold all constraints at once, we had some success in simplifying the task by reducing the number of constraints. For example, we asked GPT-4o to verify the intended bias, providing it with only the DPs and the RCs and allowing it to focus only on the relations of DP1 and DP2 to RC. %
For example, given the ambiguous-condition sentence with DP1 `schoolfriend', DP2 `preschooler' and RC `was learning to use the potty', GPT-4o flagged that ``The term `schoolfriend' suggests an older child who is likely past the potty training stage'', so we changed `schoolfriend' to `brother'. However, the check's accuracy was inconsistent. While it accepted the bias of `the neighbor of the boy who was a midwife' in EN, it rejected it in ZH, claiming that ``engaging in midwifery work can apply to any individual regardless of gender''\footnote{While male midwives do exist, a boy in this role is clearly implausible. Here, GPT-4o's debiasing mechanism may have overridden the practical implausibility of the situation.}.


\subsection{Multilingual Sets}  Starting from EN sentences, we used interactive processes with GPT-4o and Claude 3 to create initial versions in ES, JP, KO, RU, and ZH. However, this process  quickly revealed that  biases in EN do not always translate directly or maintain their relevance in other languages and cultures. As a result, we shifted from translation to adaptation, creating {language-specific sentences} that preserved the intended semantic relationships and ambiguities while respecting the linguistic and cultural norms of each language (see Appendix \ref{sec:language_specific}).

\subsection{Dataset Validation}
In order to review and adjust the sentences to ensure accuracy, cultural appropriateness, and preservation of the intended ambiguities, native-speaking professional teachers, translators, and researchers volunteered or were hired, including one of the authors, a professional translator but non-native speaker of several languages. 
This process often resulted in sentences that diverged significantly from the original EN versions, tailored to each language's specific linguistic and cultural context.

\section{Experiment 1: Humans}
To directly compare LLM and human performance in RC ambiguity resolution, we conducted a forced-choice experiment with human subjects using the EN dataset.

\subsection{Participants and Procedure} 
Sixty-five native EN speakers (mean age: 31 years; age range: 18–80 years) were recruited through the online platform Prolific.   
Participation was restricted to individuals whose first language was EN and who were residing in the United States at the time of the experiment. No participants reported a clinical history of hearing or auditory processing issues, reading difficulties, or prior brain surgery.
Participants were compensated at a rate of 15 USD per hour, upon successful completion of the task.

We conducted a forced-choice experiment using web-based survey platform PCIbex Farm \citep{zehr2018penncontroller}. 
As in example (\ref{eg:human_experiment}), after each sentence, participants were presented with a comprehension question and response options probing their interpretation of the sentence on a separate screen. The order of response options was counterbalanced.

\begin{exe}
\ex
    \noindent I saw the daughter of the woman who bought a dress.\\[1 pt]
    \hspace{1em} `Who bought a dress?'\\[1 pt]
    \hspace{1em} 1. the daughter 2. the woman
\label{eg:human_experiment}
\end{exe}

In this experiment, we tested 96 sets of EN items. These target items were distributed across three groups using a Latin square design, and fillers were included to maintain balance in the experimental conditions. The entire experiment took approximately 35 minutes to complete on average. The study protocol was approved by the Institutional Review Board (IRB).

\subsection{Analysis}
Following established psycholinguistic norms and previous studies on relative clause attachment ambiguities, an HA (non-local attachment) response rate above 0.5 is interpreted as a preference for HA, while a rate below 0.5 indicates a preference for LA (local attachment). Our analysis treated ambiguous and unambiguous cases separately. 
In the case of ambiguous sentences, we explored which attachment preference was favored, assessing whether there was a consistent inclination toward one attachment choice over the other. {In the forced-choice task, participants were presented with two response options: low attachment (LA) and high attachment (HA). Consequently, the response rates for LA and HA sum to 1. In our analysis, responses were coded as binary values, with 0 representing LA and 1 representing HA. The HA response rate was calculated directly, and the LA response rate was obtained as 1-HA rate.} This allowed us to validate the observed LA preference in EN, which aligns with established findings from earlier psycholinguistic research \citep{cuetos1988cross,mitchell2012reading}. For the unambiguous cases, we examined whether the participants’ responses corresponded with the intended biases (DP1 or DP2), ensuring the cues were correctly followed. It is important to note that human responses confirmed that biases generally led to the intended interpretation—either DP1 or DP2 attachment—in all but one of the sets.  

\subsection{Results and Discussion}\label{sec:human_result}

In ambiguous conditions, the HA response rate — calculated as the number of HA choices divided by the total number of responses — was 0.268. This rate, significantly below the 0.5 threshold, reaffirms the LA  preference in EN, consistent with findings reported in previous studies \citep{cuetos1988cross, gilboy1995argument, frazier1997construal}. When the participants' answers matched the expected responses for DP1- and DP2-biased conditions, the bias-aligned answer rates revealed a clear difference. In the DP1-biased condition, the answer rate was 0.641, indicating participants generally aligned with the DP1 bias, but with noticeable variability (sd = 0.47). In contrast, the DP2-biased condition showed a much higher answer rate of 0.885, reflecting a stronger and more consistent adherence (sd = 0.31) to the intended DP2 bias.

Our results indicate a clear LA preference in ambiguous conditions, while in unambiguous cases, humans effectively use world knowledge and biases to resolve structural ambiguities. However, discrepancies emerged depending on the targeted DP position (DP1 or DP2). Discrepancies based on DP position (DP1 or DP2) likely stem from a natural tendency to adopt LA structures in "DP1 of DP2 RC" constructions, even when world knowledge favors HA. This default to LA aligns with prior research suggesting humans favor syntactic simplicity in initial parsing \citep[a.o.]{fodor1978parsing,meng2000ungrammaticality}.

Our findings further illustrate humans' flexibility in adjusting interpretations when faced with world knowledge inconsistencies, even in rare or unconventional scenarios. For instance, in the sentence ``\textit{The doctor met the son of the woman who had a beard},'' humans still showed an LA preference, interpreting the relative clause as modifying ``\textit{the woman}'' despite the unusual nature of a woman having a beard. This flexibility can be attributed to humans' ability to adaptively incorporate context and pragmatics when processing ambiguous sentences, enabling them to accommodate infrequent but plausible real-world scenarios. The cognitive mechanism known as good-enough parsing \citep{ferreira2007good} suggests that humans do not always strive for fully accurate interpretations but rather settle for interpretations that are sufficient for comprehension, even if they require adjusting expectations based on uncommon world knowledge. This adaptability may also be linked to the fact that humans draw upon a vast reservoir of experiences and cultural knowledge, allowing them to entertain even improbable interpretations when syntactic ambiguity arises, thus showcasing their unique capacity for flexible language processing.

\section{Experiment 2: LLMs}

\subsection{Large Language Models and Procedure}\label{subsec:procedure}
We tested three widely used LLMs, using both proprietary models such as \textbf{Claude 3.5 Sonnet} (\texttt{claude-3-5-sonnet-20240620}) and \textbf{GPT-4o} (\texttt{gpt-4o-2024-05-13}), and open-source models such as the instruction-tuned \textbf{Llama 3.1} (\texttt{Meta-Llama-3.1-70B-Instruct}).


We conducted a forced-choice experiment, similar to Experiment 1, consisting of three components: sentence, question, and answer choices.
The prompt used for asking the response is \texttt{``Answer with a single number, 1 or 2, without commentary''}. This was translated into each target language (see Appendix \ref{sec:prompts_appendix} for the full prompt texts), with the respective version used for each language. The experiment was repeated three times, with each iteration varying the order of the presented choices. We tested three different configurations:

\begin{itemize} 
\itemsep-0.4em 
    \item \textbf{Linear} order: The choices were presented in the same sequence as they appeared in the sentence, e.g., in postnominal languages such as EN, this consistently meant starting with DP1, while in prenominal languages, such as KO, it corresponded to DP2.
    \item \textbf{Reversed} order: The choices were presented in the opposite sequence to how they appeared in the original sentence. 
    \item \textbf{Random} order: The choices were presented in a randomized sequence, with the randomization kept consistent across all test items and LLMs to ensure comparability. 
\end{itemize}



Prior research indicates that the order in which options are presented can affect  responses in both humans and computational models \cite{tversky1981framing, smith2007cognitive,zheng2023judging}. The use of multiple configurations aimed to control potential order effects that might influence the LLMs' responses.
 
 
\subsection{Analysis}
In our analysis, responses were examined separately for ambiguous and unambiguous cases to capture potential differences in model behavior. For ambiguous cases, we investigated the model's preference between the two possible interpretations, assessing whether there was a consistent tendency. 
For unambiguous cases, we evaluated whether the model's responses aligned with the intended bias (DP1 or DP2). Only responses selecting the provided answer choices were analyzed.

The prompt in all queries comprised a sentence, a question, two possible answers, and a request for a response consisting of only the number 1 or 2. The language-specific requests are shown below.

\begin{figure}[h!]
    \centering
\includegraphics[width=\linewidth]{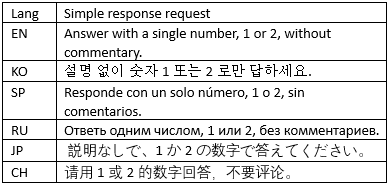}
\end{figure}

GPT-4o and Claude 3 responded consistently with 1 or 2, as requested while all outliers came from Llama-3.1, indicating slightly more variability, especially in ZH and KO. In rare cases, Llama-3.1 responded with texts (e.g., "\begin{CJK}{UTF8}{mj} 여동생\end{CJK}" [younger sister] or "\begin{CJK}{UTF8}{mj}答案： 1. 娃娃\end{CJK}" [answer: 1. baby]).  Of 15,552 responses, seven (0.045\%) were failures where Llama-3.1 chose the main subject or gave invalid responses (e.g., "-1"). 
These instances occurred across different presentation orders and languages: two outliers in the linear-order presentation (from ES), two in the reversed order (1 from ZH, 1 from KO), and three in the random-order presentation (1 from JP, 1 from ES, 1 from KO). 

To examine potential differences in LLMs' behavior across languages, we conducted a statistical analysis using mixed-effects logistic regression (see Appendix \ref{app:statistical}). All model results are reported as an average over three runs.



\subsection{Results and Discussion}\label{sec:llm_result}
\subsubsection{Human \emph{vs.} LLMs in English}
First, we compare the performance of humans and  LLMs on two aspects: (1) HA response rates in ambiguous conditions, and (2) matched answer rates for DP1- and DP2-biased conditions.  


Table \ref{tab:ambiguous_comparison} presents the HA response rates in ambiguous conditions. We observe that both humans and LLMs demonstrate a strong preference for LA ({\textgreater} 0.70), with humans exhibiting a slightly higher rate of HA responses compared to the LLMs.

Table \ref{tab:dp_comparison} presents the matched answer rates for DP1- and DP2-biased conditions, where the sentence bias promotes attachment to either DP1 or DP2. In these unambiguous cases, all models performed exceptionally well, particularly in the DP2-biased condition, where they demonstrated near-perfect accuracy and outperformed human participants. For the DP1-biased condition, GPT-4o (0.743) and Claude (0.726) still outperformed human participants (0.641), indicating a stronger alignment with the provided bias. Llama 3.1 exhibited the lowest matched answer rate (0.575) for DP1-biased sentences.\footnote{We thank the anonymous reviewer for suggesting an additional analysis. To examine how much variance in human responses can be explained by LLM responses, we conducted further analysis. Please see Appendix \ref{sec:app_human_llmprediction} for details.}


While LLMs demonstrate a robust ability to integrate world knowledge and explicit biases, they often exhibit rigidity in their interpretations, frequently reinforcing existing social stereotypes. In contrast, human participants display a notable flexibility, adapting their interpretations to align with evolving social norms and contextual subtleties. 

\setlength{\tabcolsep}{4pt}
\begin{table}[!t]
\centering
\small
\begin{tabular}{ccccc}
\toprule
\textbf{Congruency} & \textbf{Human} & \textbf{Claude 3} & \textbf{GPT4o} & \textbf{Llama 3.1}  \\ \midrule
ambiguous  & 0.268 & 0.154 
 & 0.197 & 0.157\\ 
\bottomrule
\end{tabular}
\caption{Comparison of human and model performance on HA answer rates in ambiguous conditions in EN.}
\label{tab:ambiguous_comparison}
\end{table}

\setlength{\tabcolsep}{4pt}
\begin{table}[!t]
\centering
\small
\begin{tabular}{ccccc}
\toprule
\textbf{Congruency} & \textbf{Human} & \textbf{Claude 3}  & \textbf{GPT4o} & \textbf{Llama 3.1}\\ \midrule
DP1  & 0.641 & 0.726 & 0.743 & 0.575 \\
DP2  & 0.885 & 1.000 &0.978 & 0.954 \\ \bottomrule
\end{tabular}
\caption{Comparison of human and model matched answer rates for DP1 and DP2 biased conditions in EN.}
\label{tab:dp_comparison}
\end{table}

\subsubsection{Multilingual Syntactic Attachment Preferences in Ambiguous Cases}

\begin{figure}[t!]
\centering
\includegraphics[width=0.99\linewidth]{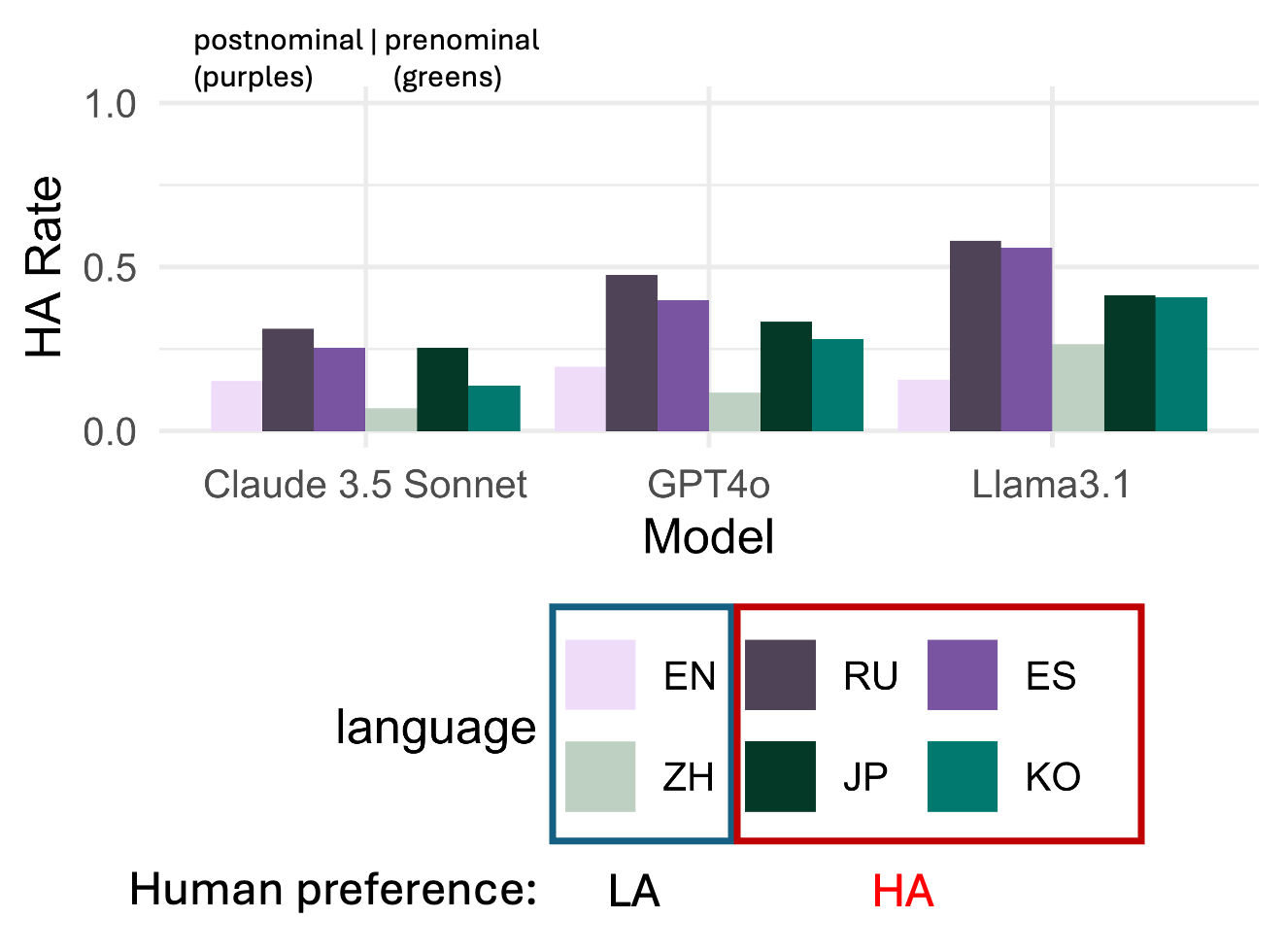} 
\caption{High attachment (HA) response rates in \textbf{ambiguous} conditions (Attachment Preference)}
\label{fig:ambiguous}
\end{figure}

Next, we examine the HA  \textit{vs.} LA  preferences of LLMs under ambiguous conditions in six languages. The results are summarized in Figure \ref{fig:ambiguous}. According to previous psycholinguistic studies, EN and ZH speakers demonstrated an LA preference, whereas JP, KO, RU, and ES speakers displayed an HA preference \citep[a.o.]{cuetos1988cross,Lee2018attachment,lee2021effect,mitchell2012reading,sekerina2003late,shen2006late}. In contrast, in our study, LLMs exhibited an overall \textbf{LA preference across all languages}, regardless of the attachment tendencies reported in previous psycholinguistic studies (detailed results in Table~\ref{tab:stat-ambi} in Appendix~\ref{app:statistical}). In addition to not reflecting language-specific preferences, the models also do not exhibit a specific pattern based on syntactic structures, such as the difference between prenominal and postnominal RC languages. 

This general tendency towards LA suggests that LLMs could be defaulting to simpler syntactic structures when resolving ambiguities rather than adapting to language-specific syntactic rules. One possible explanation is that the models may not have fully adapted to the linguistic structure of these high-attachment languages or it may be over-relying on its training data or on an innate bias learned from predominantly low-attachment languages, like EN. The one notable exception is Llama 3.1, which exhibited a slightly more HA preference in Russian, aligning with the psycholinguistic findings that often report an HA bias in human processing for this language.

\begin{figure}[t!]
    \centering
    \begin{subfigure}{0.49\textwidth}
        \centering
        \includegraphics[width=\textwidth,trim=0 60 0 0,clip]{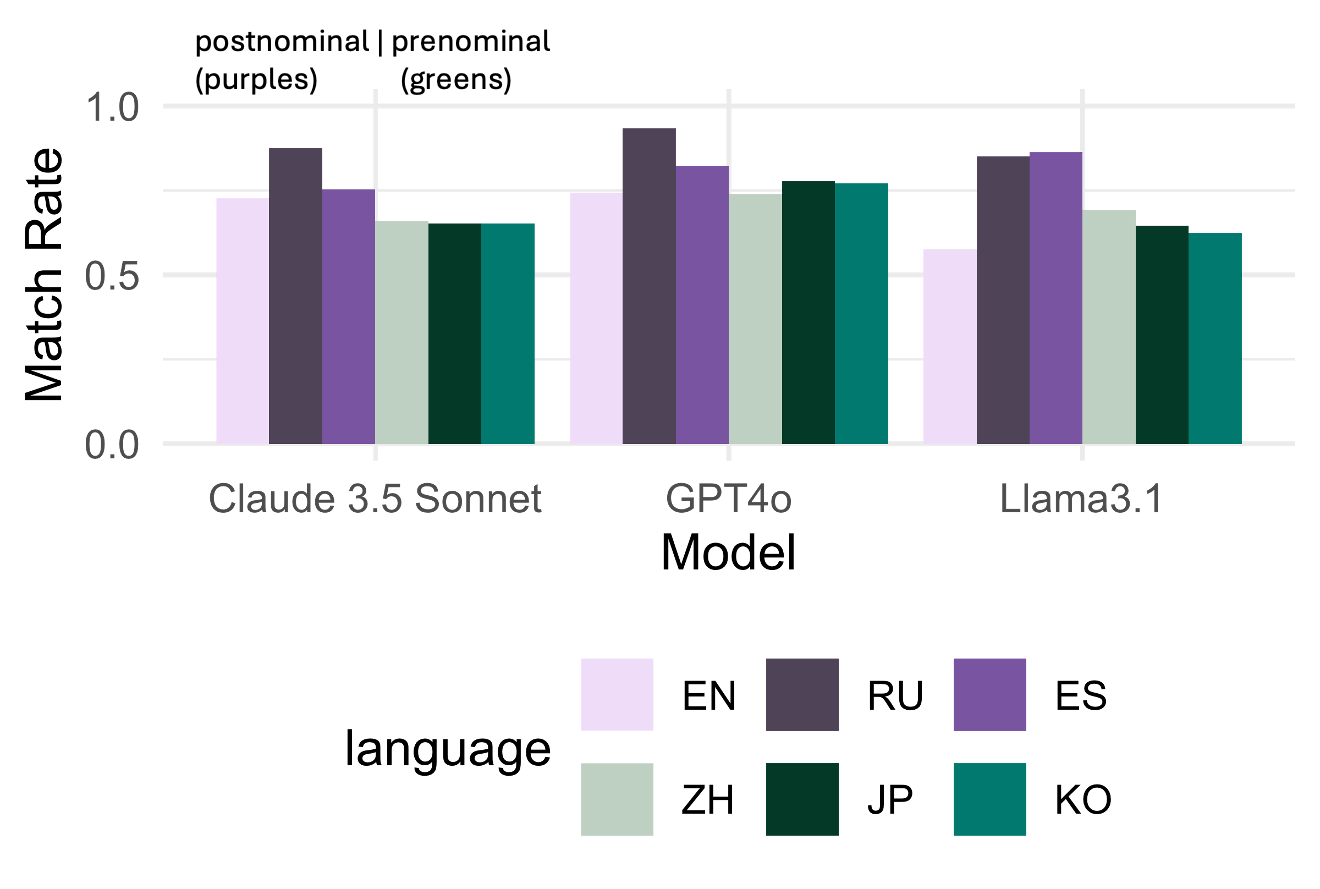}
        \caption{DP1 (high attachment)}\label{fig:sembiasresults_DP1}
    \end{subfigure}
    \hfill
    \begin{subfigure}{0.49\textwidth}
        \centering
\includegraphics[width=\textwidth]{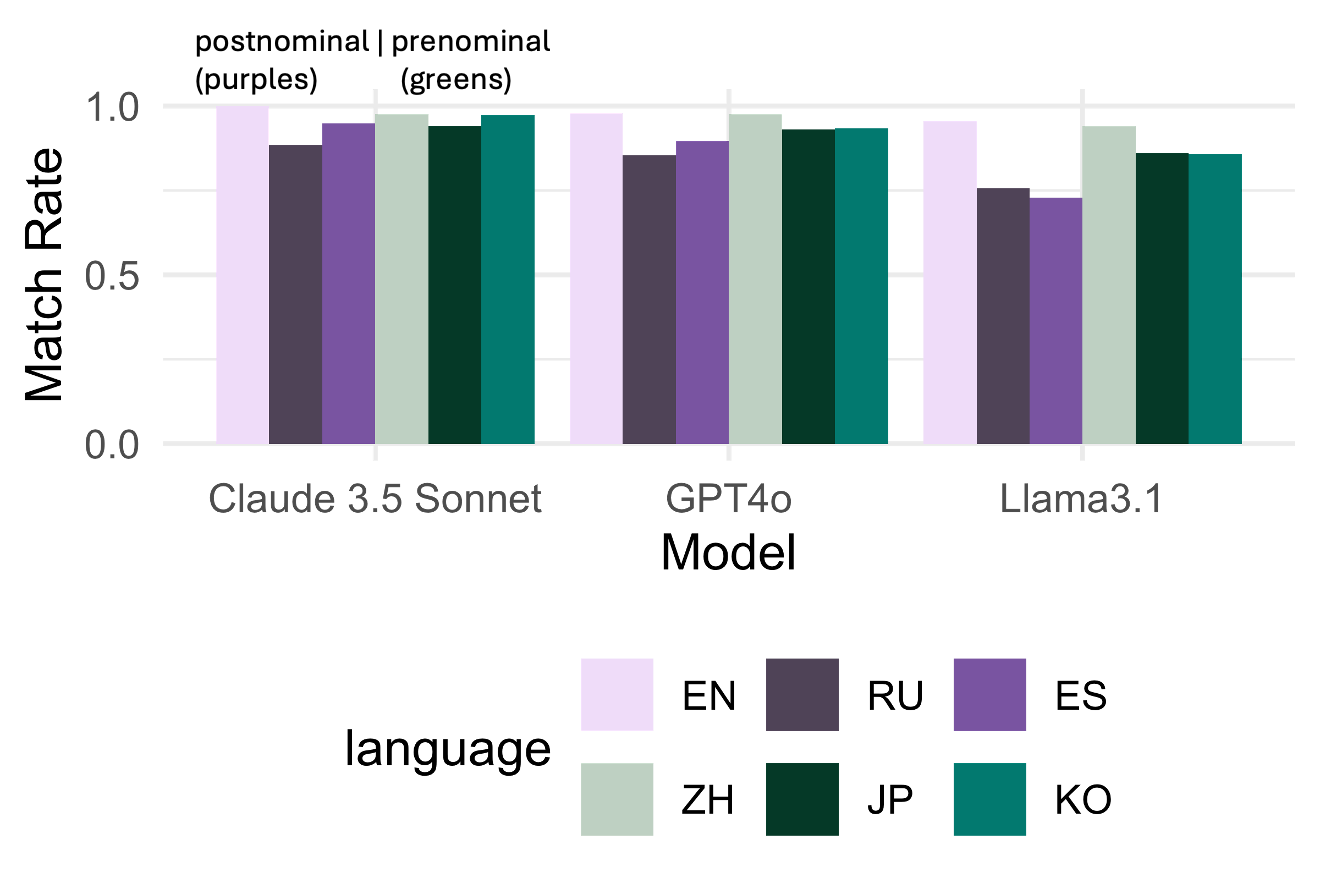}
        \caption{DP2 (low attachment)}\label{fig:sembiasresults_DP2}
    \end{subfigure}
    \caption{Average matched responses with the given world knowledge and bias toward DP1 and DP2 (high and low attachment) in \textbf{unambiguous} conditions in English }\label{fig:combined_sembiasresults}
    
\end{figure}

\subsubsection{Multilingual World Knowledge and Bias Alignment in Unambiguous Cases}
Figures \ref{fig:sembiasresults_DP1} and \ref{fig:sembiasresults_DP2} show that under unambiguous conditions, LLMs across various languages exhibit high alignment with world knowledge and bias cues.
This suggests that LLMs show strong sensitivity to world knowledge, effectively incorporating it when explicit biases are present, achieving high accuracy in interpreting syntactically unambiguous sentences (see Appendix \ref{app:details}). However, similar to humans in EN, there was a more pronounced alignment with the intended biases in DP2-biased conditions compared to DP1-biased conditions ($p$ close to 0), with Llama 3.1 showing exceptions in Russian and Spanish.



\subsubsection{Influence of Presentation Order of Answer Choices}
We analyzed the results separately based on presentation order. The models generally responded consistently regardless of presentation order of response choices, under both ambiguous (Table~\ref{tab:order_ambi})  and unambiguous cases (Appendix~\ref{app:present}). In the ambiguous cases, the models consistently showed very strong preference towards LA response rather than the ordering of answer choices, suggesting that the model is prioritizing syntactic parsing strategies over surface-level presentation biases. The model’s bias toward low attachment may lead to poorer performance in languages with high attachment preferences, particularly in tasks involving syntactic parsing, translation, or question answering. Further research is needed to ensure that the models properly adapt to the syntactic preferences of different languages.


\setlength{\tabcolsep}{7pt}
\begin{table}[t]
\small
\centering
\begin{tabular}{ccrrr}
\toprule
\textbf{Language} &    \textbf{Model} &  \textbf{linear} &  \textbf{reverse} &  \textbf{random}\\
\midrule
      \rowcolor{gray!20}EN &   Claude 3 &   \textbf{0.084} &    \textbf{0.232} &   \textbf{0.147} \\
      \rowcolor{gray!20}ZH &   Claude 3 &   \textbf{0.083} &    \textbf{0.042} &   \textbf{0.083} \\
      
      JP &   Claude 3 &   0.375 &    0.094 &   0.292 \\
      KO &   Claude 3 &   0.219 &    0.031 &   0.167 \\
      RU &   Claude 3 &   0.240 &    0.396 &   0.302 \\
      ES &   Claude 3 &   0.146 &    0.375 &   0.240 \\
      \midrule
      \rowcolor{gray!20} EN  &    GPT-4o &   \textbf{0.147} &    \textbf{0.253} &   \textbf{0.189} \\
      \rowcolor{gray!20} ZH &    GPT-4o &   \textbf{0.115} &    \textbf{0.083} &   \textbf{0.156} \\
      
      JP &    GPT-4o &   0.281 &    0.396 &   0.323 \\
      KO &    GPT-4o &   0.302 &    0.281 &   0.260 \\
      {RU} &    GPT-4o &   0.385 &    \textbf{0.552} &   0.490 \\
      ES &    GPT-4o &   0.323 &    0.479 &   0.396 \\
      \midrule
      \rowcolor{gray!20}EN & Llama-3.1 &   \textbf{0.105} &    \textbf{0.200} &   \textbf{0.168} \\
      \rowcolor{gray!20}ZH & Llama-3.1 &   \textbf{0.292} &    \textbf{0.229} &   \textbf{0.271} \\
      
      JP & Llama-3.1 &   0.458 &    0.354 &   0.432 \\
      {KO} & Llama-3.1 &   0.229 &    \textbf{0.552} &   0.442 \\
      RU & Llama-3.1 &   \textbf{0.573} &    \textbf{0.594} &   \textbf{0.573} \\
      {ES} & Llama-3.1 &   0.495 &    \textbf{0.625} &   \textbf{0.558} \\

\bottomrule

\end{tabular}\caption{Attachment preferences under varying response orders. The grayed rows (EN and ZH) indicate where humans typically show LA preference whereas all other languages show HA preferences. The cells in bold indicate where the human and LLM preferences align.}\label{tab:order_ambi}
\end{table}


\section{General Discussion}
The findings in Sections~\ref{sec:human_result} and~\ref{sec:llm_result} can be summarized as follows: While LLMs like GPT-4o and Claude 3 Sonnet exhibit high accuracy in unambiguous conditions by effectively leveraging world knowledge to resolve syntactic ambiguities, this success may reflect a reliance on stereotypes and explicit biases embedded in their training data \citep{bolukbasi2016man,sheng2019woman, lucy2021gender}, leading to rigidity in interpretations and reinforcing existing biases. In contrast, humans demonstrate greater flexibility by overriding syntactic biases when world knowledge conflicts with their default attachment preferences, considering rare but plausible interpretations through pragmatic reasoning, context, and social norms. Moreover, the LLMs' overall preference for LA indicates insensitivity to syntactic differences between prenominal and postnominal structures and an inability to adapt to language-specific attachment preferences.

These findings offer important insights into how LLMs and humans process syntactic ambiguities and incorporate world knowledge, carrying substantial implications for both the advancement of LLM technology and the understanding of human language processing. While LLMs are becoming increasingly sophisticated, they do not yet fully capture the intricate interplay of syntactic, semantic, and pragmatic factors that characterize human language processing. Human comprehension involves dynamic cognitive strategies that adapt to evolving social norms, contextual cues, and world knowledge. Integrating insights from psycholinguistics and cognitive science is essential for developing LLMs that replicate not just outputs but also the underlying cognitive mechanisms of human language processing. Future models would benefit from this integration, enabling them to handle ambiguities and culturally specific knowledge in a manner that reflects genuine human adaptability and flexibility.

\section{Related Work}

\noindent \textbf{Syntactic Ambiguity} \quad 
Early efforts used classifiers to reflect human preferences in pronoun resolution by leveraging coreference features \cite{seminck-amsili-2017-computational}. 
Other studies focused on RC attachment preferences, revealing mixed outcomes. For example, English RNNs learned low attachment preferences but over-generalized, while Spanish RNNs struggled to learn human-typical high attachment preferences unless biases in the training data were manipulated to be balanced \cite{davis2020recurrent}. In contrast, the 
\texttt{MultiWho} sentences are designed to always be 
semantically coherent, full sentences, because they 
must be realistic objects of human annotation. BERT-based parsers performed poorly on Dutch RC ambiguity unless bias-correcting priming was applied \cite{wijnholds2023structuralambiguitydisambiguationlanguage}. LLMs have been shown to diverge from human-like prepositional phrase attachment preferences \cite{cai2024largelanguagemodelsresemble}.




\medskip
\noindent \textbf{Syntactic Ambiguity Datasets} \quad A tradeoff exists between small, high-quality datasets and large, synthetic ones. For instance, BLiMP \cite{warstadt-etal-2020-blimp-benchmark}, a set of syntactic minimal pairs generated automatically from templates, has inspired similar datasets in other languages like JBLiMP \cite{someya-oseki-2023-jblimp} (a Japanese dataset of 331 minimal pairs based directly on example sentences taken from theoretical linguistics publications) and CLiMP \cite{xiang-etal-2021-climp} (a Chinese version generated synthetically from BLiMP translations, though it faced quality issues \cite{song-etal-2022-sling}, illustrating the limitations of automatic, template-based generation and translation of linguistic test sets). SLING \cite{song-etal-2022-sling} addressed these difficulties by utilizing a TreeBank to extract lexically diverse and ecologically valid sentences. 

\medskip
\noindent \textbf{Prompt Design \& Survey Bias} \quad LLMs are sensitive to minor prompt changes \citep{zhao2021calibrateuseimprovingfewshot}, with survey question modifications shifting model responses away from known human biases \citep{tjuatja2024llms}. %
Additionally, order bias (or position bias) can be introduced by the order of possible answers 
\citep{wang2023largelanguagemodelsfair,zheng2023judgingllmasajudgemtbenchchatbot,herr2024largelanguagemodelsstrategic,zheng2023judging,shi2024judging,li2023generativejudgeevaluatingalignment,liusie-etal-2024-llm}. To mitigate order bias,  \texttt{MultiWho} experiments systematically examine all answer order cases to assess the impact of order on the results. 




\section{Conclusion}
In summary, while LLMs have made considerable strides in handling syntactic ambiguities and incorporating world knowledge, they still exhibit limitations compared to human language processing. Addressing these gaps will require a concerted effort to train LLMs with more diverse and nuanced data, enhancing their adaptability and integrating insights from human cognition to create models capable of truly human-like language comprehension. Only by incorporating more context-sensitive, pragmatically rich, and culturally diverse training data can we develop LLMs that approach the depth and flexibility of human language understanding. By doing so, we will not only advance the capabilities of LLMs but also gain deeper insights into the complexities of human language processing, bridging the gap between artificial and human intelligence.


\section*{Limitations}
\label{sec:limitations}
\citet{sinha2022languagemodelacceptabilityjudgements} point out that language models' syntactic acceptability judgments improve in accuracy with longer context. Our experiments contain only a single sentence, and it is possible that longer context would improve the models' alignment with human judgment.

We sought to create  an interdisciplinary team to ground our research well in both linguistics and natural language processing. Along the way, we discovered several hurdles involved in this type of work. We, as linguists and computer scientists, speak different languages\footnote{Though we also speak several different natural languages, that's not quite what we mean.}. We look at data differently, use different terms, and consider different outcomes significant in different ways. Each of these differences posed a barrier, and overcoming these barriers grew our capabilities as interdisciplinary researchers. We have learned much from each other, and would encourage others to take on similar challenges.

\section*{Acknowledgments}
We are thankful to the anonymous reviewers for their helpful feedback.
\clearpage

\bibliography{custom}


\appendix

\section{Cross-Linguistic Adaptations and Challenges}
\label{sec:language_specific}

Creating a multilingual dataset for syntactic ambiguity resolution posed numerous challenges due to the diverse linguistic and cultural features of the target languages. This section highlights key adaptations and considerations encountered while adapting our dataset from EN to ES, JP, KO, RU, and ZH.

\subsection*{Grammatical Considerations}

In Spanish and Russian, both languages with grammatical gender, we ensured that relative clauses remained syntactically gender-neutral to maintain the required ambiguity. For example, in Spanish, "pregnant" was rephrased in one set as "in a state of pregnancy" (\textit{quedó en estado de embarazo}, \#94-96\footnote{Numbers in this section refer to the index numbers of the data items in the \texttt{MultiWho} dataset}) to avoid gender-specific adjectives. We carefully constructed sentences to avoid adjectives that modify the RC subject, which would normally provide a syntactic clue about the attachment. In Russian, we avoided past-tense verbs when one DP is feminine and the other is masculine, as past-tense verbs are gender-marked in Russian and would provide a syntactic disambiguation cue. For professions typically associated with a specific gender, we used gender-neutral alternatives. For instance, "wet nurse" in "the daughter of the man who became a wet nurse" (\#415-417) was translated to lactation professional (\textit{profesional de lactancia}, a gender-neutral term in Spanish; \#127-129).

\subsection*{Cultural Adaptations}

{Localization in the form of }cultural adaptations were necessary to ensure the stimuli were appropriate and meaningful in each language context. "Choirmaster" sounds anomalous when translated into Chinese, so was replaced with
\begin{CJK}{UTF8}{mj}"道长"\end{CJK}
\textit{daozhang} (Daoist Temple Master, \#88-90), aligning with local cultural contexts. In Spanish, "quinceañera" was used to create an age-differentiating relative clause, leveraging a culture-specific celebration (\#1324-1326). The English "tooth fairy" (\#442-444) was adapted to "ratoncito" in Spanish (\#1306-1308) and to
\begin{CJK}{UTF8}{mj}"红包"\end{CJK}
\textit{hongbao} (red envelope for gifting money to children at festivals) in Chinese (\#151-153), reflecting culture-specific concepts. Names were also adapted to be culturally appropriate, changing "John" to "Juan" in Spanish (\#1162, etc.) or "Taro" in Japanese (\#640, etc.), for example.

\subsection*{Language-specific Phraseology}

Each language presented unique challenges in expressing certain concepts while maintaining ambiguity. Succinct idiomatic expressions like "to father a child" (sets \#406-408) required creative adaptations, such as \textit{"que embarazó a una mujer"} (who got a woman pregnant) in Spanish (\#1270-1272). Register contrasts, exemplified by "The student met the preschooler of the boss who was learning to use the potty" (set \#452), posed translation difficulties. Fixed expressions like the Japanese
\begin{CJK}{UTF8}{mj}"妻帯せずに過ごした"\end{CJK}
(\#757-759), meaning "was a bachelor" but literally "spent time without entering the state of having a wife," needed careful handling to preserve semantic richness and structural ambiguity. These challenges highlight the importance of understanding each language's preferred phraseology.

The above adaptations highlight the complexity of creating equivalent stimuli across languages, and make clear that adaptations taking careful account of grammatical, cultural, and pragmatic equivalence are more appropriate than literal translations.

\section{Exploring Variance in Human Responses via LLM Predictions}
\label{sec:app_human_llmprediction}

We conducted additional analyses to explore how much variance in human responses can be explained by LLM responses.

To assess the relationship between LLM predictions and human behavior, we analyzed items (N=18) where all three LLMs provided incongruent responses. These items were then categorized based on human response patterns as follows:

Highly incongruent: 10–18 participants agreed on the incongruent answer (11/18 items).
Moderately incongruent: 1–9 participants agreed on the incongruent answer (6/18 items).
No incongruence: No participants provided the incongruent answer (1/18 item).
The alignment between LLM and humans regarding incongruency suggests LLMs can help identify ambiguous or challenging stimuli likely to result in variable human responses. This can make LLMs a useful tool for psycholinguistic research, particularly for selecting stimuli for further testing.

\setlength{\tabcolsep}{1pt}
\begin{table}[t!]
\centering
\small
\begin{tabular}{lrrrr}
\toprule
 & \textbf{Estimate} & \textbf{Std. Error} & \textbf{z value} & \textbf{Pr(>|z|)} \\
\midrule
(Intercept)    & -2.04959  & 0.15231   & -13.456  & < 2e-16 \\
languageEN     &  0.16083  & 0.14084   &   1.142  & 0.253 \\
languageJP     &  1.23326  & 0.12964   &   9.513  & < 2e-16 \\
languageKO     &  0.89965  & 0.13161   &   6.836  & 8.16e-12 \\
languageRU     &  1.86667  & 0.12861   &  14.514  & < 2e-16 \\
languageSP     &  1.60480  & 0.12869   &  12.470  & < 2e-16 \\
ModelLlama3.1  &  0.54073  & 0.08088   &   6.685  & 2.30e-11 \\
ModelSonnet    & -0.68742  & 0.08857   &  -7.761  & 8.41e-15\\
\bottomrule
\end{tabular}
\caption{Statistical Results for Ambiguous Conditions }\label{tab:stat-ambi}
\end{table}

\setlength{\tabcolsep}{1pt}
\begin{table}[t!]
\centering
\small
\begin{tabular}{lrrrr}
\toprule
 & \textbf{Estimate} & \textbf{Std. Error} & \textbf{z value} & \textbf{Pr(>|z|)} \\
\midrule
(Intercept)    &  1.41169  & 0.14735   &  9.580  & < 2e-16\\
languageEN     & -0.10699  & 0.11431   & -0.936  & 0.349      \\
languageJP     & -0.02991  & 0.11442   & -0.261  & 0.794      \\
languageKO     & -0.08231  & 0.11404   & -0.722  & 0.470     \\
languageRU     &  1.42448  & 0.13989   & 10.183  & < 2e-16 \\
languageSP     &  0.75876  & 0.12423   &  6.107  & 1.01e-09\\
ModelLlama3.1  & -0.59088  & 0.08815   & -6.703  & 2.04e-11\\
ModelSonnet    & -0.52534  & 0.08844   & -5.940  & 2.84e-09\\
\bottomrule
\end{tabular}
\caption{Statistical Results for DP1-biased Conditions}\label{tab:stat-dp1}
\end{table}

\setlength{\tabcolsep}{1pt}
\begin{table}[t!]
\centering
\small
\begin{tabular}{lrrrr}
\toprule
 & \textbf{Estimate} & \textbf{Std. Error} & \textbf{z value} & \textbf{Pr(>|z|)} \\
\midrule
(Intercept)     &  4.3175  & 0.2586   & 16.693  & < 2e-16 \\
languageEN      &  0.5305  & 0.2985   &  1.778  & 0.075473  \\
languageJP      & -1.0867  & 0.2255   & -4.819  & 1.44e-06\\
languageKO      & -0.9333  & 0.2291   & -4.074  & 4.63e-05\\
languageRU      & -1.9654  & 0.2133   & -9.216  & < 2e-16\\
languageSP      & -1.7224  & 0.2156   & -7.991  & 1.34e-15\\
ModelLlama3.1   & -1.0036  & 0.1244   & -8.067  & 7.18e-16\\
ModelSonnet     &  0.5292  & 0.1539   &  3.440  & 0.000582\\
\bottomrule
\end{tabular}
\caption{Statistical Results for DP2-biased Conditions}\label{tab:stat-dp2}
\end{table}

\section{Statistical Results}\label{app:statistical}
We conducted a statistical analysis using mixed-effects logistic regression. The primary model included LLMs and languages as fixed effects, while random intercepts were assigned to items. Following best practices, we initially employed the maximal random effects structure and progressively simplified it until model convergence was achieved \citep{barr2013random}. The analysis produced coefficients, standard errors, Z-scores, and p-values for each fixed effect and interaction, with statistical significance determined at a threshold of 0.05.

Tables~\ref{tab:stat-ambi}, \ref{tab:stat-dp1}, and \ref{tab:stat-dp2} present the statistical results for ambiguous, DP1-biased, and DP2-biased conditions, respectively.

\section{More Details in Human Results}\label{app:details}
The summary of the human results indicates that, among 96 sets, human participants provided incongruent answers in 91 sets. The number of participants per item ranged from 1 to 18. For example, in Set 2, the sentence was:
\begin{quote}
  \emph{"Mr. Johnson visited the baby of the mother who was in a stroller."}

\textbf{Question}: Who was in a stroller?

\textbf{Expected Answer}: The baby.  
\end{quote}

In Set 2, for instance, the expected answer "the baby" aligns with world knowledge and typical bias, as it is more plausible for a baby to be in a stroller than for a mother. This frequency-driven plausibility makes "the baby" the favored interpretation for the question "Who was in a stroller?" However, human participants frequently chose the incongruent response "the mother," likely due to a preference for local attachment, prioritizing the noun phrase closer to the relative clause. These incongruent human responses reflect a flexibility in interpretation that overrides world knowledge and bias in certain contexts.

\section{Presentation Order Results}\label{app:present}
Table~\ref{tab:order-dp1dp2} presents the detailed results obtained using different presentation of responses - linear, reversed, and random.

\setlength{\tabcolsep}{2pt}
\begin{table}[!t]
\small
\begin{tabular}{cccrrr}
\hline
language &    Model & congruency & linear & reverse & random \\
\hline
EN & GPT4o   & DP1 & \textbf{0.663} & 0.831 & 0.736 \\
ZN & GPT4o   & DP1 & 0.802 & \textbf{0.677} & 0.739 \\
JP & GPT4o   & DP1 & 0.791 & 0.770 & 0.770 \\
KO & GPT4o   & DP1 & 0.781 & 0.739 & 0.791 \\
RU & GPT4o   & DP1 & 0.937 & 0.937 & 0.927 \\
ES & GPT4o   & DP1 & 0.760 & 0.906 & 0.802 \\
\hline
EN & Llama3.1 & DP1 & \textbf{0.526} & 0.631 & 0.568 \\
ZN & Llama3.1 & DP1 & 0.718 & \textbf{0.642} & 0.718 \\

JP & Llama3.1 & DP1 & 0.708 & \textbf{0.614} & \textbf{0.614} \\
KO & Llama3.1 & DP1 & \textbf{0.479} & 0.770 & \textbf{0.625} \\
RU & Llama3.1 & DP1 & 0.843 & 0.864 & 0.843 \\
ES & Llama3.1 & DP1 & 0.810 & 0.916 & 0.864 \\
\hline
EN & Sonnet   & DP1 & \textbf{0.694} & 0.757 & 0.726 \\
ZN & Sonnet   & DP1 & 0.760 & \textbf{0.562} & 0.656 \\

JP & Sonnet   & DP1 & 0.760 & 0.520 & 0.677 \\
KO & Sonnet   & DP1 & 0.739 & \textbf{0.562 }& \textbf{0.656} \\
RU & Sonnet   & DP1 & 0.843 & 0.906 & 0.875 \\
ES & Sonnet   & DP1 & \textbf{0.697} & 0.812 & 0.750 \\
\hline \hline  
CH & GPT4o   & DP2 & 0.989 & 0.968 & 0.968 \\
EN & GPT4o   & DP2 & 0.989 & 0.968 & 0.978 \\
JP & GPT4o   & DP2 & 0.947 & 0.916 & 0.927 \\
KO & GPT4o   & DP2 & 0.927 & 0.937 & 0.937 \\
RU & GPT4o   & DP2 & 0.864 & 0.833 & 0.864 \\
SP & GPT4o   & DP2 & 0.947 & 0.843 & 0.895 \\
\hline
CH & Llama3.1 & DP2 & 0.916 & 0.947 & 0.958 \\
EN & Llama3.1 & DP2 & 0.968 & 0.936 & 0.957 \\
JP & Llama3.1 & DP2 & 0.885 & 0.854 & 0.843 \\
KO & Llama3.1 & DP2 & 0.916 & 0.757 & 0.895 \\
RU & Llama3.1 & DP2 & 0.802 & 0.708 & 0.760 \\
SP & Llama3.1 & DP2 & 0.812 & \textbf{0.677} & \textbf{0.697} \\
\hline
CH & Sonnet   & DP2 & 0.979 & 0.979 & 0.968 \\
EN & Sonnet   & DP2 & 1.000 & 1.000 & 1.000 \\
JP & Sonnet   & DP2 & 0.916 & 0.968 & 0.937 \\
KO & Sonnet   & DP2 & 0.968 & 0.979 & 0.968 \\
RU & Sonnet   & DP2 & 0.906 & 0.864 & 0.885 \\
SP & Sonnet   & DP2 & 0.958 & 0.937 & 0.947 \\
\hline
\label{tab:order_dp1dp2}
\end{tabular}\caption{Matched answer rates for DP1- and DP2-biased conditions obtained using different response orders}\label{tab:order-dp1dp2}
\end{table}

\end{document}